\documentclass{article}

\usepackage[final]{neurips_2025}

\usepackage[utf8]{inputenc} % allow utf-8 input
\usepackage[T1]{fontenc}    % use 8-bit T1 fonts
\usepackage{hyperref}       % hyperlinks
\usepackage{url}            % simple URL typesetting
\usepackage{booktabs}       % professional-quality tables
\usepackage{amsfonts}       % blackboard math symbols
\usepackage{algorithm}
\usepackage{algpseudocode}
\usepackage{enumitem}
\usepackage{wrapfig}
\usepackage{amsmath}
\usepackage{amssymb}
\usepackage{bm}
\usepackage{svg}
\usepackage{nicefrac}       % compact symbols for 1/2, etc.
\usepackage{microtype}      % microtypography
\usepackage{xcolor}         % colors
\usepackage{multirow}
\usepackage{graphicx}
\usepackage{subcaption}
\usepackage{cleveref}
\usepackage{fancyvrb}
\usepackage{xcolor}
\usepackage{tikz}
\usetikzlibrary{arrows.meta, positioning}
\usepackage{algorithm}
\usepackage{algpseudocode}
\usepackage{amsmath}
\usepackage{pgfplots}
\pgfplotsset{compat=1.18}
\usepackage{float}
\usepgfplotslibrary{groupplots}
\usepgfplotslibrary{statistics}
\usepackage{tabularx}
\usepackage{booktabs}
\usepackage{tablefootnote}
\usepackage{array}
\usepackage{threeparttable}
\usepackage{placeins}
\usepackage{longtable}

\title{
  Symphony for Text Generation\\[0.5em]
  \large Benchmarking Clinical Note Generation
}
\author{%
    \small
    \textbf{Daniel Varab}
    \And
    \small
    \textbf{Victor Petrén Bach Hansen}
    \And
    \small
    \textbf{Matthew Newton}
    \And
    \small
    \textbf{Asbjørn W Helge} 
    \And
    \small
    \textbf{Kevin Pelgrims}
    \And
    \small
    \textbf{Mathias Baltzersen*}
    \And
    \small
    \textbf{Adrian Young-San Roessler}
    \And
    \small
    \textbf{Vanessa Klungtvedt}
    \And
    \small
    \textbf{Maximilian Brand}
    \And
    \small
    \textbf{Lasse Krogsbøll}
    \And
    \small
    \textbf{Henrik Cullen}
    \And
    \small
    \textbf{Lars Maaløe}
}

\begin{document}
\maketitle

\begin{abstract}
Ambient documentation systems are rapidly gaining adoption, yet their impact on clinical note quality remains poorly characterized. We introduce MedConv, a multilingual dataset of 300 clinical encounters in English, Danish, and German, and use it alongside the Ambient Clinical Intelligence benchmark (ACI-BENCH) to compare Corti, a clinical AI platform, with two leading, accessible ambient scribe software applications built on general-purpose AI. We present a controlled clinical evaluation framework that combines entailment metrics with LLM-judged pairwise comparisons across eight dimensions adopted from PDSQI-9. Results show that Corti's API-based text-generation infrastructure is on par with or outperforms leading commercial scribes. We further show that Corti's configurable API provides the flexibility necessary to fine-tune quality dimensions for specific documentation use cases. We present the evaluation methodology and release a dataset to support future reproducible comparison of ambient documentation systems.

\end{abstract}

\begin{figure}[H]
  \centering
  \makebox[\linewidth][c]{%
      \includegraphics[
        width=1.20\linewidth,
        height=6.5cm,
        keepaspectratio,
        trim={2cm 0.3cm 0.5cm 0},
        clip]{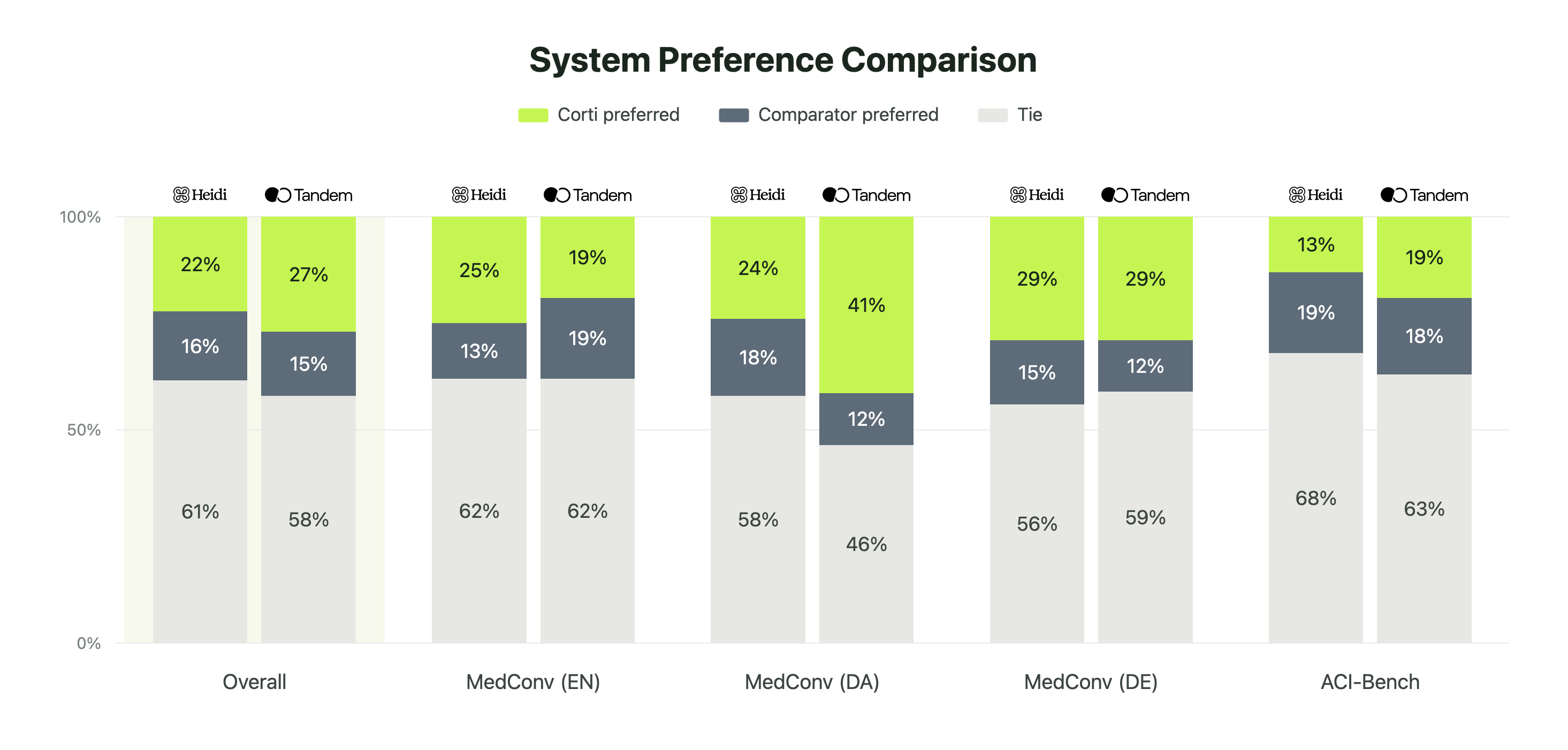}%
  }
  \caption{Pairwise preference outcomes on SOAP notes, shown for each dataset and overall pooled. For each Corti--comparator pair, bars report the percentages of judgments preferring Corti, preferring the comparison system, or resulting in a tie, aggregated over the eight adapted PDSQI dimensions. Contradictory judgments are counted as ties. The pooled result comprises 3,296 dimension-level judgments per compared system (412 encounters $\times$ 8 dimensions).}
  \label{fig:pairwise-outcomes-soap}
\end{figure}

\section{Introduction}

Ambient documentation systems are increasingly used to reduce clinical documentation burden and recent studies report improvements in clinician experience, well-being, and documentation efficiency across a range of deployment settings~\citep{vanlinschoten2026ambient,Olson2025Burnout,Pearlman2025Efficiency,Lukac2025RCT,Afshar2025RCT}. As the adoption and workflow benefits improve, the quality of the clinical documentation remains largely under-reported. 

Clinical note quality is difficult to compare systematically across ambient documentation systems: existing cross-vendor studies vary in scale and in how tightly they control the inputs and configurations each system receives~\citep{anderson2025ambient,Draper2025Safety,fox2026onenoteinthree}. Omissions as a dominant, critical error type with potentially severe patient impact, are especially hard to measure, since evaluators that inspect only what a note contains can miss what it left out~\citep{Draper2025Safety,schumacher2025medomit, fox2026omissionblindness}.

Surface-level measures, such as BLEU \citep{papineni2002bleu}, METEOR \citep{banerjee2005meteor} and ROUGE \citep{lin2004rouge}, are poorly suited to clinical text (\S\ref{sec:pointwise}), and recent approaches, therefore, combine semantic or factual assessment, clinician review, structured quality instruments, and LLM-based evaluation to resolve their shortcomings~\citep{wang2025scribe,croxford2025pdsqi,croxford2025llmjudges}.

As a result, teams building ambient scribes (or health systems trying to select one) are forced to rely on qualitative surveys and anecdotal evidence. This creates the risk of conflating the performance of underlying AI platform with factors that are extraneous to it, such as workflow integration, user interface design, or the preferences of a lead clinician on a pilot team.

In this report, we present a methodology and a public dataset for evaluating ambient documentation performance. We demonstrate how a purpose-built AI platform matches or outperforms commercial ambient scribe software applications built on general-purpose AI platforms while enabling simple, safe tailoring to clinician preferences. The goal is to provide anyone building AI solutions for clinical documentation a way to easily tailor documentation towards specific workflow needs and quantify the resulting performance.

Our contributions are as follows.
\begin{itemize}
    \item \textbf{A multi-dimensional evaluation protocol for clinical note generation.} We score each note on three automatic entailment measures: groundedness (\textit{is it supported by the consultation?}), completeness (\textit{does it include what the reference note includes?}), and conciseness (\textit{does it include grounded content  that's omitted from the reference?}). We also use LLM judges to compare notes from two systems side-by-side using validated clinical criteria adopted from PDSQI-9, with clinicians further validating a subset of the judgments for correctness.

    \item \textbf{A multilingual clinical documentation dataset.} We introduce MedConv, da dataset comprising of 300 diverse synthetic clinical scenarios with matched transcript--note pairs in English, Danish, and German. Designed to support controlled evaluation across clinical domains and encounter settings.

    \item \textbf{A controlled multilingual cross-system benchmark.} We evaluate two widely adopted ambient documentation applications built on general purpose AI platforms, Heidi and Tandem Health, and the vertical AI platform Symphony by Corti on the MedConv and ACI-BENCH datasets.

    \item \textbf{An empirical analysis of template configurability.} We show that Corti's configurable AI platform enables small, targeted changes in prompt design to meaningfully shift quantitative performance, allowing clinicians to tailor output quality along the dimensions most relevant to their use case.

\end{itemize}

\clearpage

\section{Methodology}

\subsection{Experimental Design}\label{sec:experimental-design}
We design a paired benchmark to compare clinical note quality and evaluate three candidate systems. To focus the comparison on note generation, we consider only text transcripts as input, excluding potential speech recognition errors. Systems generate notes from identical pre-transcribed inputs under a harmonised four-section SOAP template, using default generation settings for each system. All evaluation measures are computed at encounter level. SOAP is the only note type evaluated, being the format most consistently supported across the three systems and three markets. Notes are assessed through complementary evaluation, measuring both content fidelity and fine-grained documentation quality. We consider both reference-based, and reference-free metrics to measure coverage of notes, as well as paired comparisons to assess the relative quality of notes generated from the same encounter across domain specific quality dimensions.

\subsection{Evaluation Datasets}\label{sec:datasets}

To compare systems we use two datasets: the publicly available ACI-BENCH corpus \citep{yim2023aci}, and MedConv, a dataset created by Corti that extends coverage to additional medical domains, consultation settings, and languages (see \autoref{tab:dataset-stats}).

\begin{table}[H]
    \centering
    \caption{Dataset sizes, mean transcript length, and estimated durations at 150\,words-per-minute.}
    \label{tab:dataset-stats}
    \renewcommand{\arraystretch}{1.1}
    
    \begin{tabular}{lrrc}
    \toprule
    Dataset & $N$ & \#words (min--max) & Avg. encounter duration (min) \\
    \midrule
    ACI-BENCH (EN) & 112 & $1271 $ (710--2541) & 8.5 \\
    MedConv (DA)   & 100 & $1655 $ (923--2585) & 11.0 \\
    MedConv (DE)   & 100 & $1822 $ (904--2739) & 12.1 \\
    MedConv (EN)   & 100 & $2031 $ (930--3073) & 13.5 \\
    \bottomrule
    \end{tabular}
\end{table}

\paragraph{ACI-BENCH} The Ambient Clinical Intelligence Benchmark (ACI-BENCH) corpus, published by Microsoft Health AI, Nuance Communications, and University of Washington, comprises of 207 cases across three subsets representing different modes of clinical note generation from doctor-patient conversations\citep{yim2023aci}. We use the \textit{aci} subset, since it most closely matches the ambient-scribe scenario evaluated in this benchmark. While the dataset does not come with explicit medical domain labels, LLM-assisted analysis points to diverse coverage across 10 domains, with musculoskeletal/orthopedic dominating and cardiovascular, respiratory, gastrointestinal, and hepatology well represented. Note structures are diverse; the original authors grouped them into four standardized meta-divisions, Subjective, Objective-exam, Objective-results, and Assessment-plan, which align closely with the SOAP structure adopted for this work.

\paragraph{MedConv} The Corti MedConv corpus comprises of 100 transcript-notes pairs per language (English, Danish, German; 300 cases total) spanning 15+ medical specialities. 100 English notes are generated by seeding a synthetic pipeline with clinician authored stories and clinical clues. Each clinical note is validated and corrected by the same clinicians. Transcripts are generated synthetically using another synthetic pipeline which conditions on the validated clinical note, constructing a noisy multi-speaker consultation transcript. The process is repeated for each target languages. Encounter settings are mixed: outpatient (32), inpatient (24), admission (19), general practice (11), pre-hospital (7), and investigation (7). Reference note structures are diverse and tailored to individual cases and domains, and are not generated according to Corti's templates.\footnote{The MedConv dataset will be released publicly alongside a paper detailing its construction and validation. Upon release, the dataset will be available at \url{https://huggingface.co/datasets/corti/med-conv}.}

\clearpage

\subsection{Systems Under Review}
\paragraph{Heidi and Tandem Health}
Heidi\footnote{\url{https//www.heidihealth.com/}} and Tandem Health\footnote{\url{https://www.tandemhealth.ai/}} are stand-alone clinician-facing ambient documentation software applications, supporting template-configured clinical note generation. We select these clinical systems, rather than general-purpose model APIs, such as OpenAI's GPT models or Google's Gemini model family, as they represent some of the fastest growing, industry-leading software application integrations of such models in healthcare contexts\footnote{Other commercial scribes were not considered due to access limitations or limited language coverage.}. The challenges identified in these well-resourced software applications are likely to be even greater for developers of software applications with fewer resources. 

\paragraph{Corti}
Corti’s Symphony offers a clinical AI platform that includes purpose-built capabilities for speech-to-text \citep{nix2026symphonyspeechtotextsupportingrealtime}, text generation, medical coding \citep{edin2026symphonymedicalcodingnextgeneration}, agents, and general model inference\footnote{\url{https://docs.corti.ai}}. Its text-generation capabilities provide an API-based alternative to using general-purpose LLMs directly, supporting transcript- and/or fact-based document synthesis with configurable, section-level templates and structured outputs. For this evaluation, Corti was accessed through Symphony’s text-generation APIs.\footnote{\url{https://docs.corti.ai/textgen}}

For a full reference of the configurations used for each system, see \S\ref{appendix:system-configs} in the appendix.

\paragraph{Note Generation}
Notes are generated in September 2026 using each system’s available interface: Corti via the Text Generation API\footnote{See \S\ref{sec:corti-text-gen} in the appendix for a detailed description of the full pipeline}, and Heidi and Tandem Health manually through their end-user web interface - providing transcripts as consultation context and exporting generated notes as plain text. For Corti, we sample five notes per encounter across all four datasets through the API. Because Heidi and Tandem Health require manual generation through web interfaces, it was intractable to run large-scale comparisons. As a result, five notes per encounter where sampled for each system on ACI-BENCH, and one per encounter on the MedConv datasets. This allows measuring variation across systems while recognising that variation may not generalise to MedConv. Within each system, inputs, templates, and configurations, are fixed across replicates, such that between-run variation reflects only stochastic nature of the generation pipelines.

\paragraph{Configuration and template harmonization} We consider one canonical SOAP note type per system and geography, harmonizing headings as far as each template editor allowed. Changes are limited to section reordering, heading renames, and splitting or merging sections; we do not rewrite vendor prompt text (\autoref{tab:heidi-soap}, \autoref{tab:Tandem Health-soap}). We record output-affecting settings and fix them to vendor defaults, with Corti in its default production configuration (\autoref{tab:system-configurations}). Market configurations werare UK English, German, and Danish; the US-origin ACI-BENCH cases are evaluated under the UK configuration.

\subsection{Evaluation Measures}

\paragraph{Metrics} Surface-level measures such as BLEU, ROUGE, and METEOR assume that overlap in wording implies proximity in semantics. Clinical documentation violates in both directions: equivalent findings are routinely expressed with disjoint vocabulary, and a negation or reversed causal relation can render a note false while leaving nearly all wording intact. Because errors of the latter kind are lexically conservative, an $n$-gram scorer will rank a clinically incorrect note above an acceptable paraphrase, which makes these measures least reliable on precisely the failure modes that carry the greatest clinical risk.

For this reason, similar to \citet{hansen2025factsr}, we adopt a model-as-a-judge approach when evaluating the quality of the generated notes. An LLM is used to assess the quality of generated outputs in place of traditional automatic metrics or human annotators \citep{gao2024llmbasednlgevaluationcurrent,li2024llmsasjudgescomprehensivesurveyllmbased,yehudai2025surveyevaluationllmbasedagents}.

Given the multi-faceted complexity of evaluating clinical note generation, we propose a multi-dimensional evaluation framework built on two complementary judgments. The first is reference-based: does the generated note capture the clinically relevant content of the reference note without adding content beyond it, and is everything it states supported by the transcript?
We operationalize this as three entailment-based metrics adapted from \citet{hansen2025factsr}, in which an LLM judge checks whether each statement in one text is supported by another: reference statements against the generated note (completeness), generated statements against the reference (conciseness), and generated statements against the transcript (groundedness).

Entailment scores capture content fidelity but not the dimensions clinicians weigh when judging nuanced note quality. PDSQI-9, introduced by \citet{croxford2025pdsqi, croxford2025llmjudges}, is a clinician-validated instrument to measure the quality of clinical notes in a set of fixed dimensions (a.o. organization, synthesis, succinctness, and stigmatizing language). We extend the instrument to a preference-based measure and compare notes pairwise across eight of the instruments dimensions. To reduce position bias, we evaluated each note pair in both orders. We assign system preference only when both judgments agree, and otherwise record a tie (\S\ref{sec:pairwise}). We validate the extension against blinded clinician preferences on a targeted case subset (\S\ref{appendix:judge-validation}) and include document judge-model sensitivity of the measure in (\S\ref{appendix:judge-sensitivity}). We provides a preference measure that allows direct evaluation of note-generation pipelines and, as we show in \S\ref{sec:template-configuration}, can be used to monitor improvements and regressions related to scribe or template configuration changes.

For detailed definitions we refer to the appendix in \S\ref{sec:pointwise} for the entailment-based metrics and \S\ref{sec:pairwise} for the pairwise-adapted PDSQI-9 dimensions.

\paragraph{Latency}
Runtime was recorded for each generated note: API-side for Corti, from request submission to receipt of the complete generated document, and browser-side for comparators, from submission to the fully rendered note. Measurements were taken sequentially from Copenhagen in early September 2026.

\paragraph{Statistical analysis}
Pairwise preference scores are reported with 95\% confidence intervals from a non-parametric cluster bootstrap over encounters (\S\ref{sec:pairwise}); intervals including 50 indicate no detectable preference. We report all eight dimensions without multiplicity correction, so dimension-level results are descriptive rather than independent hypothesis tests. The two families' uncertainty estimates are not interchangeable: one captures generation variability without encounter sampling, the other the reverse.

\paragraph{Choice of judge}
Pairwise preferences depend on the judge model and are sensitive on the chosen models biases. We assess sensitivity by computing comparisons with OpenAI's GPT-5.4 and Anthropic's Opus 4.6 (\S\ref{appendix:judge-sensitivity}). We observe similar preference distributions, however, with Opus 4.6 generally assigning more ties. We selected GPT-5.4 as the primary judge for all model-based metrics as it yielded fewer indeterminate comparisons.

\subsection{Clinical Validation of the Pairwise Evaluation}\label{sec:clinical-validation}

We further employ human evaluation to validate preference measures. We task three in-house clinicians, with diverse backgrounds (German, Norwegian, Danish) across multiple specialities, to validate the LLM verdicts on each of the PDSQI-dimensions. 10 notes from the ACI-BENCH runs are sampled and a blinded pairwise comparison of the notes generated by two systems are presented in a simple web-based user interface to each clinician separately. See Figure \ref{fig:pdsqi-validation-ui-1} and \ref{fig:pdsqi-validation-ui-2} in the appendix for the interfaces presented to the clinicians. For each of the sample cases, the user interface presented a side-by-side view of the two notes, randomly flipped to avoid position bias, and the reasoning traces of both position-swapped judgments for the dimensions. 

Clinicians endorse the judge's verdict in 207 of 240 assessments (86.3\%, Gwet's AC1 = 0.694), and a majority of the three clinicians are in agreement in 75 of 80 case-dimension comparisons, supporting the validity of the pairwise judge as a scalable proxy for clinician preference (\S\ref{appendix:judge-validation}). Agreement was highest for accuracy, thoroughness, and usefulness and lowest for synthesis (AC1 = 0.080).

\clearpage
\section{Results}
\subsection{Overall performance evaluation of the three systems}

\begin{table}[!t]
\centering
\caption{Content related metrics across datasets. Highest scores within each dataset are shown in bold. $\pm$ denotes standard deviation over means across multiple runs; \emph{--} marks single-run entries.}
\label{tab:text-generation-results-soap}
\small
\setlength{\tabcolsep}{5pt}
\begin{tabular}{llccc}
\toprule
Dataset
& System
& Groundedness $\uparrow$
& Completeness $\uparrow$
& Conciseness $\uparrow$ \\
\midrule
	\multirow{3}{*}{ACI-Bench (EN)} & Corti & \textbf{96.3\% $\pm$ 0.3\%} & \textbf{77.3\% $\pm$ 0.2\%} & 74.9\% $\pm$ 0.2\% \\
	 & Heidi & 94.9\% $\pm$ 0.4\% & 75.9\% $\pm$ 0.3\% & \textbf{75.7\% $\pm$ 0.2\%} \\
	 & Tandem Health & 94.7\% $\pm$ 0.3\% & 73.1\% $\pm$ 0.4\% & 74.0\% $\pm$ 0.1\% \\
\midrule
	\multirow{3}{*}{MedConv (EN)} & Corti & \textbf{97.8\% $\pm$ 0.1\%} & \textbf{73.0\% $\pm$ 0.1\%} & 81.4\% $\pm$ 0.3\% \\
	 & Heidi & 95.4\% $\pm$ \rlap{--}\phantom{0.0\%} & 64.6\% $\pm$ \rlap{--}\phantom{0.0\%} & \textbf{83.0\% $\pm$ \rlap{--}\phantom{0.0\%}} \\
	 & Tandem Health & 96.7\% $\pm$ \rlap{--}\phantom{0.0\%} & 62.9\% $\pm$ \rlap{--}\phantom{0.0\%} & 81.3\% $\pm$ \rlap{--}\phantom{0.0\%} \\
\midrule
	\multirow{3}{*}{MedConv (DA)} & Corti & \textbf{97.1\% $\pm$ 0.3\%} & \textbf{71.9\% $\pm$ 0.4\%} & \textbf{85.7\% $\pm$ 0.2\%} \\
	 & Heidi & 94.8\% $\pm$ \rlap{--}\phantom{0.0\%} & 64.1\% $\pm$ \rlap{--}\phantom{0.0\%} & 85.6\% $\pm$ \rlap{--}\phantom{0.0\%} \\
	 & Tandem Health & 89.7\% $\pm$ \rlap{--}\phantom{0.0\%} & 61.8\% $\pm$ \rlap{--}\phantom{0.0\%} & 78.8\% $\pm$ \rlap{--}\phantom{0.0\%} \\
\midrule
	\multirow{3}{*}{MedConv (DE)} & Corti & \textbf{98.1\% $\pm$ 0.2\%} & \textbf{74.5\% $\pm$ 0.2\%} & 82.6\% $\pm$ 0.2\% \\
	 & Heidi & 93.9\% $\pm$ \rlap{--}\phantom{0.0\%} & 65.1\% $\pm$ \rlap{--}\phantom{0.0\%} & \textbf{84.1\% $\pm$ \rlap{--}\phantom{0.0\%}} \\
	 & Tandem Health & 96.7\% $\pm$ \rlap{--}\phantom{0.0\%} & 65.3\% $\pm$ \rlap{--}\phantom{0.0\%} & 82.9\% $\pm$ \rlap{--}\phantom{0.0\%} \\
\bottomrule
\end{tabular}
\end{table}

Corti produces the most complete notes on all four datasets, while the three systems score similarly on groundedness and conciseness (\autoref{tab:text-generation-results-soap}). Here, \textit{completeness} measures how much of the clinically relevant information in the reference note is captured, \textit{groundedness} measures whether statements in the generated note are supported by the source transcript, and \textit{conciseness} measures whether the generated note avoids bloat by outputting grounded content absent from the reference.

Completeness is the lowest-scoring dimension for all systems (62–77\%) and the clearest source of differentiation. Corti’s advantage ranges from 1 percentage point on ACI-BENCH to 8–10 points on MedConv, suggesting that the systems diverge more on the more challenging MedConv cases. By contrast, groundedness is consistently high (90–98\%), and both groundedness and conciseness are generally within two points across systems; Tandem Health on MedConv Danish is the main exception scoring lower. Run-to-run standard deviations are at most 0.4 points across five repetitions (for all systems on ACI-BENCH and Corti on MedConv) suggesting consistent outputs. An exploratory breakdown of the flagged statements by failure type and severity is provided in \S\ref{appendix:ungrounded-details} of the appendix.

Corti has the lowest mean latency across all datasets, at 5.9 to 9.7 seconds per note, while Heidi took 2.0 to 2.7 times as long and Tandem Health 1.9 to 2.2 times as long (\autoref{tab:latency}). 

\begin{table}[h]
\centering
\caption{Speedup ratios and dataset statistics relative to Corti.}
\label{tab:latency}

\begin{tabular}{lrrrrrr}
\toprule
 & & Corti & \multicolumn{2}{c}{Heidi} & \multicolumn{2}{c}{Tandem Health} \\
\cmidrule(lr){4-5} \cmidrule(lr){6-7}
Dataset & Transcript chars & avg (s) & avg (s) & vs Corti & avg (s) & vs Corti \\
\midrule
ACI-BENCH (EN) & 6101 (3474--12097) & 5.86 & 15.59 & 2.7$\times$ & 11.33 & 1.9$\times$ \\
MedConv (DA) & 7591 (4243--12115) & 9.68 & 19.68 & 2.0$\times$ & 21.41 & 2.2$\times$ \\
MedConv (DE) & 9551 (5269--15258) & 9.07 & 19.22 & 2.1$\times$ & 18.97 & 2.1$\times$ \\
MedConv (EN) & 8960 (4973--13949) & 7.49 & 17.60 & 2.3$\times$ & 14.18 & 1.9$\times$ \\
\bottomrule
\end{tabular}
\end{table}

\subsection{Granular evaluation on clinical dimensions}

\begin{figure}[!t]
    \centering
    \makebox[\textwidth][c]{%
        \includegraphics[width=1.2\linewidth]{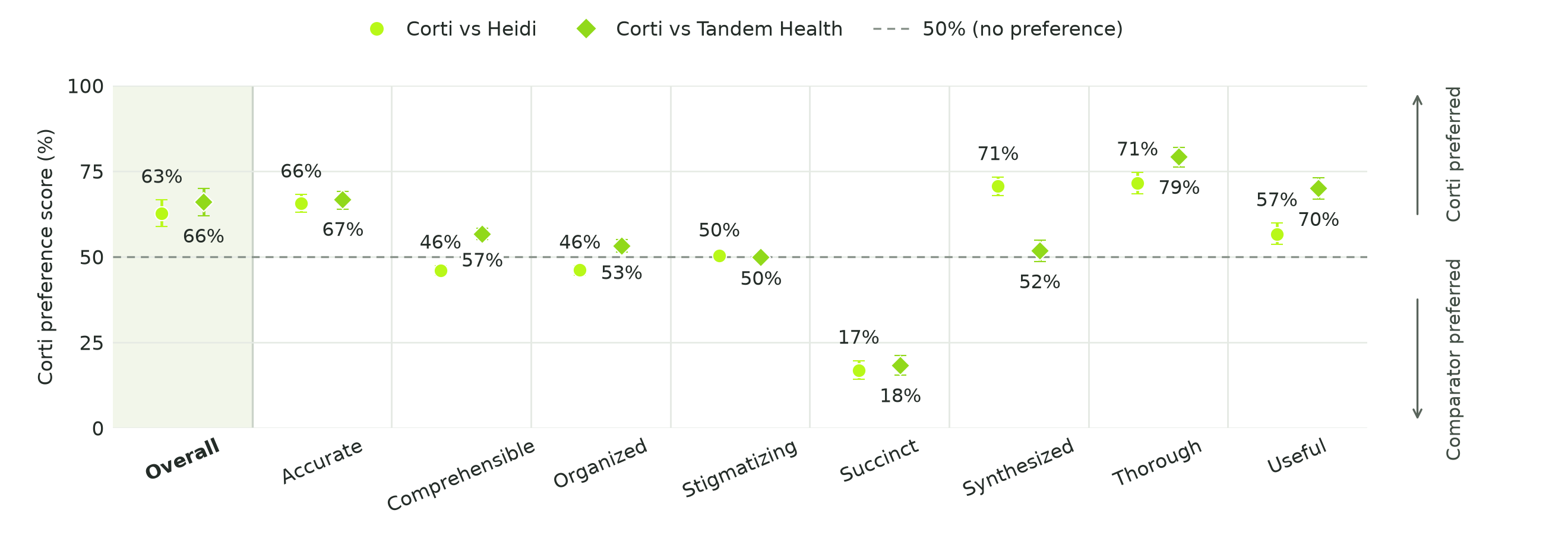}
    }
    \caption{PDSQI-9 preference scores on SOAP, comparing Corti with each evaluated system.}
    \label{fig:overall-preference}
\end{figure}

A pairwise preference evaluation provides a more granular view of where the systems differ clinically. Results are displayed in \autoref{fig:overall-preference}. Rather than reducing note quality to aggregate metrics, it compares outputs across rich dimensions such as accuracy, thoroughness, usefulness, organization, comprehensibility, synthesis, and succinctness. This reveals trade-offs that are not apparent from the entailment-focused evaluation alone: systems with similar overall groundedness or conciseness can still differ substantially in how complete, useful, well-structured, or appropriately synthesized their notes are.

Pooled across datasets, the LLM judge prefers Corti's notes over both Heidi's and Tandem Health's (\autoref{fig:pairwise-outcomes-soap}). Corti is preferred 22\% of the evaluations against Heidi at 16\% and 27\% against Tandem Health at 15\%. The only case where an ambient scribe application is preferred against Symphony is on ACI-BENCH where Heidi is preferred 19\% of the evaluations against Corti at 13\%. The high proportion of ties indicates well-calibrated and high-performing systems overall (46--68\%).

Across all dimensions, Corti wins overall 63\% and 66\% of the time against Heidi and Tandem Health respectively.\footnote{Refer to \S\ref{sec:pairwise} in the appendix for the definition of the \emph{overall} win-rate.} Against both comparators, Corti is preferred on \textbf{accuracy}, \textbf{thoroughness}, and \textbf{usefulness}, and the comparator is preferred on \textbf{succinctness} (\autoref{fig:overall-preference}). Corti's largest advantage is on thoroughness (71.5 against Heidi, 79.1 against Tandem Health) and its largest deficit is on succinctness (16.9 and 18.2). The comparator-dependent dimensions split as follows: against Heidi, Corti is preferred on synthesis (70.6) but slightly less preferred on comprehensibility and organization (both 46); against Tandem Health, Corti is preferred on comprehensibility and organization and roughly neutral on synthesis (51.7).

\subsection{Targeted intervention to improve on succinctness}\label{sec:template-configuration}  

Different documentation contexts demand different levels of granularity. To test a targeted intervention on improving the succinctness dimension for the Corti API, we test a template optimized for General Practice which equates some simple changes to the prompts in the template. We add the following general-practice instruction:
\begin{quote}
\small
\textit{This is a general-practice visit note. Keep it brief and deliberately selective: document only the clinically salient, decision-relevant information and leave out the rest. Synthesize the source into short, covering statements to achieve a correct but easily readable note.}
\end{quote}
Furthermore, we change the writing-style configuration from a comprehensive, fluent style using sentences of up to 20 words to a terse, detailed style using telegraphic statements of up to 15 words, and added compression instructions into the subjective section:
\begin{quote}
    \small
    \textit{Aim for 3–4 short sentences. The default is to OMIT}
\end{quote}
We repeat the pairwise evaluation against the comparators. See \autoref{tab:corti-soap} for more details on templates. 

\begin{figure}[t]
    \centering
    \makebox[\textwidth][c]{%
        \includegraphics[width=1.2\linewidth]{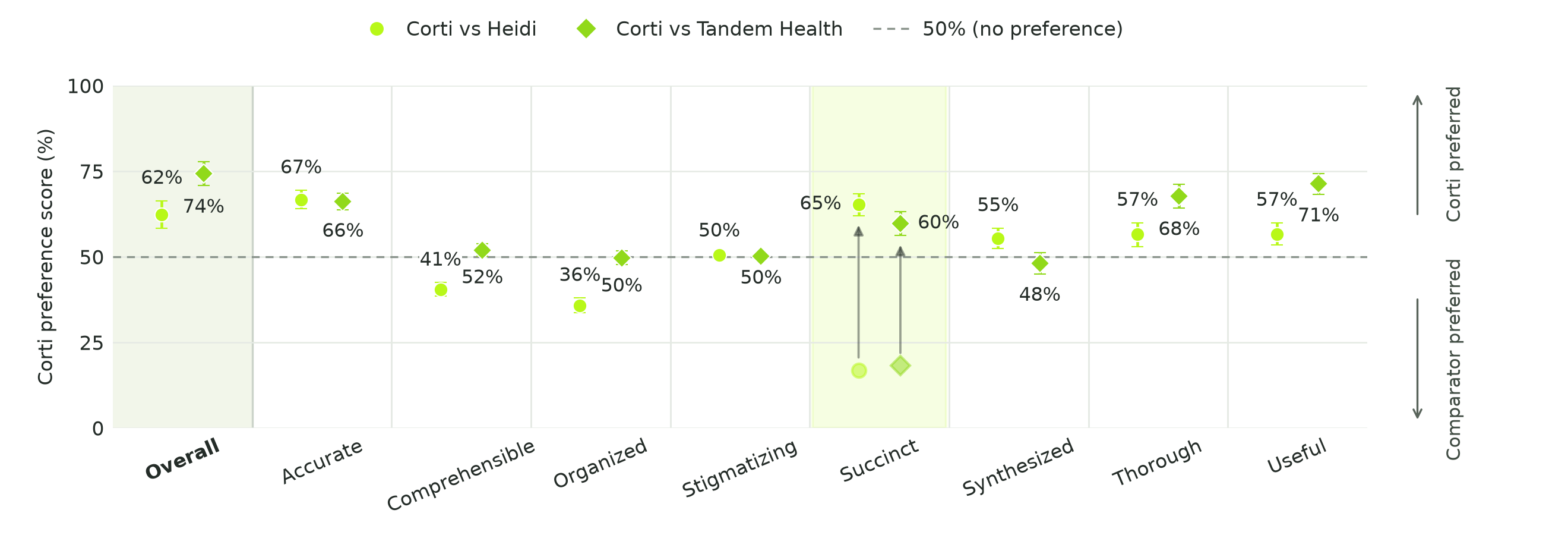}
    }
    \caption{PDSQI-9 preference scores after adding general-practice instructions, comparing Corti with each evaluated system. Each dot represents a comparison between Corti and one other system.}
    \label{fig:overall-preference-gp}
\end{figure}

\autoref{fig:overall-preference-gp} displays the preference scores under this change. The succinctness preference score increases from 16.9\% to 65.3\% against Heidi. Against Tandem Health, it increases from 18.2\% to 59.7\%, making the Corti notes the preferred option also on the succinctness dimension.

As a trade-off, thoroughness score decreases but stays favourable for Corti above 50\%. Other dimensions do not change uniformly. Against Heidi, organization decreases from 46.1\% to 35.8\% and comprehensibility from 46.0\% to 40.5\%. Against Tandem Health, synthesis changes from 51.7\% to 48.1\%. 

By modifying only two narrowly scoped instruction fields while leaving the upstream base section unchanged, the intervention illustrates the practical value of section-level over monolithic template definitions

\section{Conclusion}

To address the growing need for multifaceted evaluations of ambient clinical notes in light of patient safety concerns, we present a controlled, multilingual framework across groundedness, completeness, conciseness, and multidimensional note quality. The evaluated systems differ in completeness and pairwise quality preferences. Corti achieves the highest completeness and groundedness scores across all four datasets, at the fastest generation runtime, and is preferred over both comparators on average across the clinical dimensions of the PDSQI framework.

The relative importance of completeness and succinctness depends on the clinical context: in complex or speciality encounters, where omissions may carry greater risk, completeness may be prioritized, whereas in high-volume, lower-complexity settings clinicians may favour brevity to reduce review burden. A small, surgical template intervention shifted Corti over comparators in succinctness while preserving an above-neutral preference for thoroughness, demonstrating that documentation behaviour can be steered through API-configurations by integrators, e.g., health provider or software integrator. This highlights that ambient documentation should be evaluated not only as a fixed product, but as a configurable systems whose optimal quality profile depends on the clinical setting and workflow.

More broadly, with this study, we highlights the importance of reproducible, programmatic evaluation. By releasing MedConv and presenting our evaluation framework, we aim to make comparisons of ambient documentation systems more transparent, controlled, and repeatable, to support health systems and scribe builders in evaluating both note quality and the degree of control they have over it.

\section{Acknowledgements}
We thank Jonas Lyngsø and Fabio Voermans for the design and implementation of this text generation approach and powerful API, Chituru Chinwah and the Corti AI Platform team for assuring quality and enabling a highly performant infrastructure, Robert James for making the MedConv dataset a reality. Thank you to the rest of Corti for their valuable contributions, insightful discussions, and support throughout the development and evaluation of this work.

\newpage
\bibliography{references}

\clearpage
\appendix

\section{Related Work}
\paragraph{The role and status of ambient scribes in healthcare systems}
A growing body of research has characterized the benefits of ambient documentation systems, common challenges in piloting and implementation, and patterns of adoption across healthcare markets. Recent randomized trials and multi-site evaluations suggest that ambient scribes can reduce documentation burden and improve aspects of clinicians' well-being, although effects vary across products, users, and clinical settings \cite{Olson2025Burnout,Pearlman2025Efficiency,Lukac2025RCT,Afshar2025RCT,Chowdhury2026Crossover,Rotenstein2026Multisite}.

Evidence on the adoption and impact of ambient scribes in Europe remains more limited than in the United States. However, emerging studies from the Netherlands, Sweden, Ireland, Spain, and preliminary work from Denmark report similar benefits and challenges. These systems can reduce perceived documentation burden and improve clinician experience by freeing mental capacity and reducing perceived work pressure, while their effects on consultation time and operational efficiency remain variable \citep{vanlinschoten2026ambient, sanmark2026evaluation,Olsson2026HTA,Bracken2026Inpatient,vallejo2026escriba,DanishAmbientPsykatriPreview2026}.
Implementation studies further show that ambient scribes alter clinical communication and redistribute documentation work. Sustained use thus depends on workflow fit, local adaptation, and organisational support \cite{Cresswell2026Implementation,Seuren2026NASSS,Tierney2025Scale}. Reported challenges include the continued need for clinician review and correction, reduced performance in complex or multilingual consultations, technical and EHR integration issues, and uneven sustained adoption following initial pilots. At the organisational level, successful scaling also requires effective onboarding, workflow redesign, demonstrable financial returns, and increasingly mature approaches to data governance, clinical safety, and regulatory compliance \citep{HASCNIL2026, Lakemedelsverket2026, potts-gartner2026roi}.

\paragraph{Comparative studies of ambient scribes}
Prior US-centric studies have compared commercial ambient scribes in provider-specific implementation settings focused primarily on time-saving and stress-reduction outcomes, pointing to quality issues but without going into detailed comparisons \citep{Lukac2025RCT,Chowdhury2026Crossover}. 
Comparisons using shared encounters and direct note-quality measures report substantial variability. \citet{anderson2025ambient} compared five ambient-scribe platforms on shared simulated encounters; \citet{fox2026onenoteinthree} audited three commercial scribes, though differences in input modality and template contributions limit what can be inferred about the generation pipelines; and \citet{Draper2025Safety} compared seven scribes on eight consultations, finding summarisation accuracy was generally high, omissions were the dominant error type, and error frequency varied widely across scribes.

\paragraph{Evaluation of Clinical Documentation Systems}
Evaluating generated clinical documentation requires assessing whether a note preserves important information, remains supported by the encounter, and presents that information usefully. Existing frameworks such as SCRIBE combine automated metrics, simulation, clinician review, and LLM-based evaluation~\citep{wang2025scribe,Taylor2026NoteSafety}. Omissions are a central challenge in evaluating generated clinical notes: information missing from a note can escape evaluators that primarily inspect the statements it contains, thus omission-focused work provides a complementary perspective: MED-OMIT decomposes dialogue into facts and assesses omitted information through its relevance to a simulated differential diagnosis~\citep{schumacher2025medomit}. More recent experiments show that explicitly enumerating source facts before checking the note can improve omission detection over whole-note judgments, although detection remains imperfect and calibration does not necessarily transfer to vendor-generated notes~\citep{fox2026omissionblindness}.

Structured quality rubrics provide another complementary approach. Based on the initial Physician Documentation Quality Instrument (PDQI) introduced by \citet{Stetson2012AssessingEN} \citep{palm2025}, the Provider Documentation Summarization Quality Instrument (PDSQI-9) was developed and validated for LLM-generated summaries of clinical records~\citep{croxford2025pdsqi}. Subsequent work evaluated LLM judges against physician assessments using this instrument, supporting the potential for scalable, multidimensional assessment without establishing exhaustive clinical-error detection~\citep{croxford2025llmjudges}.

\clearpage
\section{Symphony for Text Generation} \label{sec:corti-text-gen}

\paragraph{Text Generation API endpoints}
Symphony exposes clinical text and document generation via several API endpoints spanning general LLM inference, clinical fact extraction (FactsR \citep{hansen2025factsr}) and guided document synthesis, and working in tandem with Symphony for speech-to-text \citep{nix2026symphonyspeechtotextsupportingrealtime}, Symphony for medical coding \citep{edin2026symphonymedicalcodingnextgeneration}, and agentic orchestration of multi-step clinical workflows. The guided document synthesis API accepts multiple forms of clinical context and return schema-controlled, structured clinical documents based on fully API-configurable templates. This modular architecture enables straightforward customization and orchestration of document-generation workflows, while supporting integration across any form factor. Rather than functioning as a standalone point solution, ambient documentation can be embedded directly into larger automated workflows supporting downstream actions.

\paragraph{Clinical context and fact extraction}
Text Generation supports several input representations, including conversation transcripts, extracted clinical facts, clinician notes taken during the encounter in the form of dictations or free-form clinical text, existing medical documents, and combinations of these sources. Clinical source material can be passed directly to document synthesis or first transformed into structured clinical facts using FactsR.\footnote{https://docs.corti.ai/textgen/factsr} The latter provides an intermediate representation between the source transcript from the consultation and the final note, allowing clinically relevant propositions to be identified and reviewed.

\paragraph{Guided document synthesis}
Documents are generated through guided, template-controlled synthesis. The generation layer receives the selected source context together with a template and produces a structured clinical document rather than unconstrained free-form text. Output can range from conventional narrative sections to lists, typed fields, arrays, nested objects, and other representations suitable for direct integration into downstream systems or EHR fields. 

\paragraph{Templates and sections}
Templates are first-class, UUID-addressable resources composed of reusable, typed sections with defined output schemas. Rather than representing a document template as one monolithic prompt that is difficult to debug and calibrate, each section exposes several narrower control surfaces that separately specify what information should be included or excluded, the desired voice and writing style, additional synthesis instructions, and the required output format. This decomposition separates \emph{what} should be documented, \emph{how} it should be expressed, and \emph{how} it should be structured, allowing each dimension to be isolated, tested, and tuned independently.

Sections can be reused and inherited across templates, allowing organizations to derive specialty-, department-, or use-case-specific variants from common base sections while changing only the instructions that differ. Improvements to a shared base section can therefore propagate across dependent templates without requiring equivalent changes to hundreds of independently maintained prompts. Corti additionally provides a curated library of clinically pre-configured ``Corti Standards'' templates and sections that can be combined directly, partially customized, or replaced with organization-specific definitions. Together with template versioning and publishing, this provides a governed and programmatically reproducible mechanism for controlling clinical documentation behavior. All template configuration functionality is available directly through an API and as a user-interface through the Corti Console. \footnote{https://console.corti.app/}

\paragraph{Note-generation pipeline used in these experiments}

In the experiment below, notes are generated through the full pipeline: FactsR first extracts facts from the transcript, after which both are passed to guided document synthesis together with any additional context. This reflects a typical real-time workflow in which transcripts and facts are generated during the encounter and subsequently used for note generation. Although FactsR supports clinician review, we use the extracted facts without modification to ensure a fair comparative analysis.

\section{System configurations}
\label{appendix:system-configs}

\begin{table*}[!ht]
\centering
\caption{Configuration of the three clinical note-generation systems evaluated in the benchmark.
All systems were provided with pre-transcribed consultation content; speech-to-text performance was therefore outside the scope of the comparison.}
\label{tab:system-configurations}

\small

{
\renewcommand{\arraystretch}{1.5}

\begin{tabularx}{\textwidth}{
    >{\raggedright\arraybackslash}p{3.0cm}
    >{\raggedright\arraybackslash}X
    >{\raggedright\arraybackslash}X
    >{\raggedright\arraybackslash}X}
\toprule
\textbf{Parameter} &
\textbf{Corti} &
\textbf{Heidi} &
\textbf{Tandem Health} \\
\midrule

Access method &
Text Generation API &
End-user web interface &
End-user web interface \\

Input &
Pre-transcribed consultation transcript; additional extracted clinical facts available to the generation pipeline &
Pre-transcribed consultation transcript provided as context &
Pre-transcribed consultation transcript provided as context \\

Speech-to-text evaluated &
No &
No &
No \\

Markets / languages &
UK English, German, Danish &
UK English, German, Danish &
UK English, German, Danish \\

Clinical note templates &
SOAP and GP-SOAP configurations &
SOAP configurations &
SOAP configurations \\

Template source &
Corti Standards and evaluation-specific configurations &
Heidi defaults for each market, adapted where necessary &
Tandem Health defaults where available; templates created from scratch where no suitable default was available \\

Template representation &
Typed, section-based templates with explicit section-level instructions for content, style, synthesis, and output structure &
Editable template structure with vendor-provided defaults and visible prompt configuration &
Section- and heading-based templates; some system-provided section instructions are not exposed to the evaluator \\

Template harmonization &
Section structure normalized to the common benchmark taxonomy &
Vendor templates minimally adapted to the common benchmark taxonomy &
Vendor templates minimally adapted to the common benchmark taxonomy \\

Prompt transparency &
Section-level generation instructions available to the evaluator &
Template prompt visible through the product's Structure view &
Underlying system-provided section prompts not fully visible \\

Generation configuration &
Evaluation configuration held fixed across all cases within each condition &
Scribe: Simple; Voice: Goldilocks (default) &
Assessment: ``Brief assessment \& Plan'' (default);
Level of detail: Standard (default) \\

Configuration policy &
Fixed for each experimental condition &
Vendor defaults retained except for required template normalization &
Vendor defaults retained except for required template normalization \\

Evaluation date &
September 2026 &
September 2026 &
September 2026 \\

\bottomrule
\end{tabularx}

}
\end{table*}

\FloatBarrier

\subsection{Template Configuration Details}
\begin{table}[!ht]
\centering
\caption{Heidi SOAP template configuration by country.}
\label{tab:heidi-soap}
{
\renewcommand{\arraystretch}{1.5}
\begin{tabular}{llp{8cm}}
\toprule
\textbf{Country} & \textbf{Default available} & \textbf{Adjustments} \\
\midrule
UK & Yes & Moved prompts of separate Past Medical History section into Subjective section to normalise to 4 SOAP sections\\
Germany & Yes & Adjusted German SOAP-Notiz by moving all of Past Medical History (Vergangene Krankengeschichte) under Subjective section to normalise to 4 SOAP sections. Removed (S), (O), (A), (P) from headings to normalise.\\
Denmark & Yes (English prompts) & Exact same adjustments as with UK SOAP \\
\bottomrule
\end{tabular}

}
\end{table}

\begin{table}[!ht]
\centering
\caption{Tandem Health SOAP template configuration by country.}
\label{tab:Tandem Health-soap}

{
\renewcommand{\arraystretch}{1.5}

\begin{tabular}{llp{8cm}}
\toprule
\textbf{Country} & \textbf{Default available} & \textbf{Adjustments} \\
\midrule
UK & No & Created from scratch, headings only \\
Germany & Yes & Based on standard German EHR template (SOP), separated ``Beurteilung'' (Assessment) from ``Plan'' to normalise to 4-section SOAP \\
Denmark & Yes & Based on SOAP for Almen Praksis, separated ``Vurdering'' (Assessment) from Plan to normalise to 4-section SOAP \\
\bottomrule
\end{tabular}

}
\end{table}

\begin{table}[!ht]
\centering
\caption{Corti SOAP template configurations}
\small Note: You can access Corti templates and their configurations via Corti Console or the public API to retrieve the full prompt configuration for each uuid listed below.
\label{tab:corti-soap}
{
\renewcommand{\arraystretch}{1.5}
\begin{tabular}{p{1.8cm} p{9cm} p{3cm}}
\toprule
\textbf{Country} & \textbf{Default available} & \textbf{Adjustments} \\
\midrule
UK & Yes \newline SOAP: \url{f901c06a-70db-59f6-8d0a-0a4bef4b8c77} \newline succinct SOAP : \url{b9f26972-f8ff-5a6d-aa75-e9cd43d783fa} & none\\
Germany & Yes \newline SOAP: \url{549b5f41-aed2-51fe-b0a2-f963bb7d6555} \newline succinct SOAP: \url{d85de77b-e5e9-5b5c-976a-ea21fc0e0b1c} & none\\
Denmark & Yes \newline SOAP: \url{dc601541-c241-5ee5-90ad-b39bf53516e4} \newline succinct SOAP: \url{c85d2aaa-26ee-5a81-986f-a4cbdeaed9eb} & none \\
\bottomrule
\end{tabular}

}
\end{table}

\FloatBarrier

\clearpage
\section{Template Architecture Differences}

\begin{table*}[!ht]
\centering
\begin{threeparttable}

\caption{Template architecture and configuration transparency
across the evaluated systems.}
\label{appendix:template-architecture}

\small

{
\renewcommand{\arraystretch}{1.5}

\begin{tabularx}{\textwidth}{
    >{\raggedright\arraybackslash}p{2.7cm}
    >{\raggedright\arraybackslash}X
    >{\raggedright\arraybackslash}X
    >{\raggedright\arraybackslash}X
}
\toprule
\textbf{Property} &
\textbf{Corti} &
\textbf{Heidi} &
\textbf{Tandem Health} \\
\midrule

Template representation &
Typed, section-based templates composed of reusable document sections &
Editable template structure with section headings and associated prompting &
Primarily section- and heading-based templates \\
\addlinespace

Content control &
Explicit section-level instructions define content to include or exclude &
Content instructions can be specified within the editable template prompt &
Template headings define intended content; additional custom instructions can be added \\
\addlinespace

Style and synthesis control &
Separate controls for writing style, synthesis instructions, and output representation &
Style and content instructions are expressed within the template prompt &
Custom prompting and user- or clinic-level preferences can influence generated output \\
\addlinespace

Output structure &
Section-specific, schema-controlled outputs, including typed structured fields &
Template structure determines the organization of the generated document &
Template headings determine document structure; selected fields can use structured output types \\
\addlinespace

Prompt transparency &
Section-level content, style, synthesis, and formatting instructions are exposed &
Template prompt can be inspected through the Structure view &
System-provided section-level instructions are not fully exposed \\
\addlinespace

Reuse and inheritance &
Sections are reusable and versioned; templates can inherit from shared base sections &
Vendor defaults can be copied, modified, or replaced with custom templates &
Vendor defaults can be modified or new templates can be created \\
\addlinespace

Vendor-provided templates &
Curated Corti Standards library of clinical templates and reusable sections &
Vendor-provided defaults plus a community library of clinician-created templates &
Market- and specialty-specific vendor defaults where available \\
\addlinespace

Localization &
Templates can be configured as locale-specific variants &
Market-specific defaults are available, although prompt localization varies by market &
Market-specific defaults are available, with template availability varying by market \\

\bottomrule
\end{tabularx}

}

\end{threeparttable}
\end{table*}
\FloatBarrier

\clearpage

\section{Evaluation metrics}
\subsection{Entailment-based evaluation metrics}\label{sec:pointwise}

In this work, we frame the semantic alignment metrics of \citet{hansen2025factsr} as a textual entailment task and use an LLM as the entailment judge. Given a segmented \emph{source} (the premise) and a set of segmented \emph{hypotheses}, the judge decides, for each hypothesis independently, whether it is entailed by the source. Because the judge reasons only over the sentences it is given and may not introduce outside facts, the result is a \emph{directional} semantic relation between the transcript $T$, the reference note $D_R$, and the generated note $D_G$, rather than a learned proxy of preference.

Formally, the judge takes a premise segmented into sentences $P = (p_1, \dots, p_m)$ and hypotheses $H = (h_1, \dots, h_n)$, and assigns every hypothesis $h_j$ a label
\[
\ell(h_j) \in \{\texttt{entailed},\ \texttt{partially entailed},\ \texttt{not entailed},\ \texttt{not applicable}\},
\]
citing the source sentences it relied on. A hypothesis is \texttt{entailed} when every claim it makes is stated in, or follows by deduction from, one or more cited source sentences; \texttt{partially entailed} when its main claim is supported but it adds a detail the source does not support (e.g., an unstated dose, frequency, duration, or laterality) or only some of its claims hold; \texttt{not entailed} when it is unsupported or contradicted; and \texttt{not applicable} for section headers and non-clinical boilerplate. A hypothesis may be supported jointly by several source sentences, and a source sentence may support several hypotheses.

Each label is mapped to a weight, \texttt{entailed} $=1$, \texttt{partially entailed} $=\tfrac{1}{2}$, \texttt{not entailed} $=0$, and a metric is the mean weight over the applicable hypotheses:
\[
\mathrm{score}(H) = \frac{\sum_{j\,:\,\ell(h_j)\neq\texttt{n/a}} w\big(\ell(h_j)\big)}{\big|\{\, j : \ell(h_j)\neq \texttt{n/a} \,\}\big|},
\]
where \texttt{not applicable} hypotheses are excluded from both numerator and denominator, and the score is defined as $1$ when every hypothesis is \texttt{not applicable} so that boilerplate-only text is not penalized. The intermediate \texttt{partially entailed} label makes this finer-grained than a binary supported/unsupported decision.

The three metrics differ only in which document acts as the premise and which supplies the hypotheses:
\begin{enumerate}
  \item \textbf{Completeness}: premise $D_G$, hypotheses the segments of $D_R$, giving the proportion of $D_R$ entailed by the generated note. A score of 100\% means all meaning in $D_R$ is captured; 0\% means none is.
  \item \textbf{Conciseness}: premise $D_R$, hypotheses the segments of $D_G$, giving the proportion of the generated note entailed by $D_R$, reflecting clinical relevance. A score of 100\% means all generated content is relevant; 0\% means none is.
  \item \textbf{Groundedness}: premise $T$, hypotheses the segments of $D_G$, giving the proportion of the note entailed by the transcript. This differs from a hallucination rate because unsupported sentences may also stem from entailment-judge or segmentation error rather than a genuine fabrication.
\end{enumerate}

Completeness and conciseness thereby play the roles of recall and precision of the generated note against $D_R$, while groundedness measures support in $T$.

\subsection{Pairwise Evaluation}\label{sec:pairwise}

To capture fine-grained differences in quality between systems, we evaluate relative note quality using model-based pairwise measures. An LLM compares two notes generated from the same source transcript and determines which note better satisfies a set of predefined quality dimensions. This provides a measure grounded in differences between the notes, as opposed to global calibrated score, avoiding challenges associated with numerical scores and LLM-based judges~\citep{zhou2025evaluating}. Additionally, being a reference-free measure, it provides a scalable means of evaluation on a much larger and more diverse collection of clinical input, than that by the scarce availability of annotated clinical notes.

For rubric dimensions, we adopt the validated quality measures for evaluating the quality of clinical records, \textit{the Provider Documentation Summarization Quality Instrument} (PDSQI-9)~\citep{croxford2025pdsqi, croxford2025llmjudges}. See \autoref{tab:pdsqi_dimensions} for a definition of each dimension. For each pair, the judge receives the source transcript, two notes, all dimension definitions, and returns one of three outcomes for each dimension---Note A, Note B, or Tie---together with a brief justification. The following dimensions are included: \textit{Accurate}, \textit{Thorough}, \textit{Useful}, \textit{Organized}, \textit{Comprehensible}, \textit{Succinct}, \textit{Synthesized}, and \textit{Stigmatizing}. We exclude the \textit{Cited} attribute because the evaluated systems do not produce provenance citations.

We add an aggregate \textit{Overall} dimension based on the majority vote of the eight retained dimensions. For each of the eight dimensions, the judge prefers system A, prefers system B, or records a tie. A system receives the overall preference if it is preferred on at least five dimensions; otherwise, the overall result is a tie. We recognise that note quality is highly subjective, and that there is no evidence that dimensions are linearly correlated with preference, however, this provides a simple but expressive summary statistic to guide intuition.

\begin{table}[htbp]
\centering
\caption{Dimensions of the Provider Documentation Summarization Quality Instrument (PDSQI-9) \cite{croxford2025pdsqi}.}
\label{tab:pdsqi_dimensions}
\renewcommand{\arraystretch}{1.5}
\begin{tabularx}{\textwidth}{lX}
\toprule
\textbf{Dimension} & \textbf{Description} \\
\midrule
Cited &
Claims are accompanied by appropriate citations to the source documentation. \\

Accurate &
Information is factually correct and free from fabrication, falsification, or other incorrect content. \\

Thorough &
Important clinical information is sufficiently complete, without clinically relevant omissions. \\

Useful &
Included information is relevant and useful to the clinician receiving the summary. \\

Organized &
Information is structured coherently so that the clinical course and relevant information can be followed easily. \\

Comprehensible &
The summary is clear and unambiguous and uses understandable language and terminology. \\

Succinct &
Information is communicated concisely, avoiding unnecessary repetition or redundant content. \\

Synthesized &
Information is integrated into a coherent clinical synopsis that reflects appropriate inference and medical reasoning. \\

Stigmatizing &
The summary is free from stigmatizing language about the patient. \\
\bottomrule
\end{tabularx}
\end{table}

To mitigate positional bias, a pair is evaluated twice. In the first pass, system outputs $X$ and $Y$ are presented as Note A and Note B, respectively; in the second pass, their positions are reversed. We then map both responses from positional labels back to the canonical system identities, yielding $z_1,z_2\in\{X,Y,T\}$, where $T$ denotes a tie. The final outcome is defined conservatively as
\begin{equation}
    c(z_1,z_2)= \begin{cases}
        X, & z_1=z_2=X,\\
        Y, & z_1=z_2=Y,\\
        T, & \text{otherwise}.
    \end{cases}
\end{equation}

A note is declared \textit{preferred} only if both judgments select it. Any disagreement resolves as a tie. While this prevents position-sensitive judgments from being counted as a win, it does not remove other sources of model-judge bias.

For a system under review $X$, we summarize the adjudicated outcomes using a tie-adjusted preference score
\begin{equation}
    \operatorname{PS}(X)=100\,\frac{N_X+\tfrac{1}{2}N_T}{N_X+N_Y+N_T},
\end{equation}

where $N_X$, $N_Y$, and $N_T$ are the numbers of adjudicated wins for $X$, wins for $Y$, and ties, respectively. A score of 50 is neutral; values above 50 favor $X$, and values below 50 favor $Y$.

To estimate 95\% confidence intervals for the preference scores, we use a non-parametric bootstrap over paired notes. Within each bootstrap replicate, ordered pairs are sampled with replacement, and the preference score is recomputed from the resulting win, loss, and tie outcomes. All dimension-level outcomes belonging to an encounter are resampled together, preserving their within-encounter dependence. For results pooled across datasets, resampling is performed separately within each dataset to preserve the original dataset sizes. The reported interval is defined by the 2.5th and 97.5th percentiles of the scores obtained from 10{,}000 bootstrap replicates.

\subsection{Judge Sensitivity}
\label{appendix:judge-sensitivity}
\begin{figure}[htbp]
    \centering
    \includegraphics[
        width=1\linewidth,
    ]{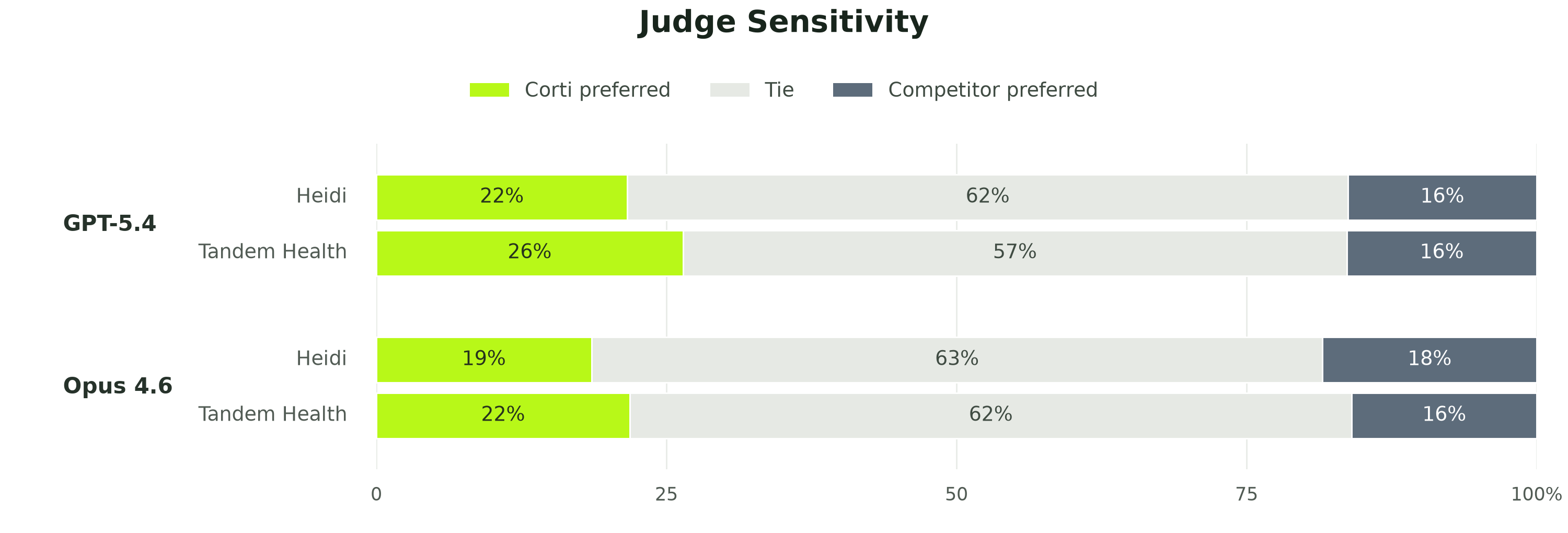}
    \caption{Judge sensitivity under GPT-5.4 and Opus 4.6 across all datasets and dimensions.}
    \label{fig:judge-sensitivity}
\end{figure}

Adopting LLM-as-a-judge measures, inherently adopts the bias and preferences of the chosen underlying judge model. We conduct a simple sensitivity analysis of preference scores to identify potential variation across LLM families. We consider OpenAI's GPT-5.4 \citep{openai2026gpt54} and Anthropic's Opus 4.6 \citep{anthropic2026opus46}. We recompute preference scores anew, as described in \S\ref{sec:pairwise}, and display "Overall" preference scores in ~\autoref{fig:judge-sensitivity}. Both models exhibit similar preferences distributions. Opus 4.6 more frequently scores pairs as ties than GPT-5.4, while both models generally lean towards Corti as being the more preferred output \textit{Overall}. The results show little difference between model families, indicating low judge sensitivity for the proposed pairwise comparison method. 

\section{System Latency Extended}
\label{appendix:latency-extended}
\begin{figure}[htbp]
\centering
\begin{minipage}{0.85\linewidth}
    \centering
    \includegraphics[width=0.48\linewidth]{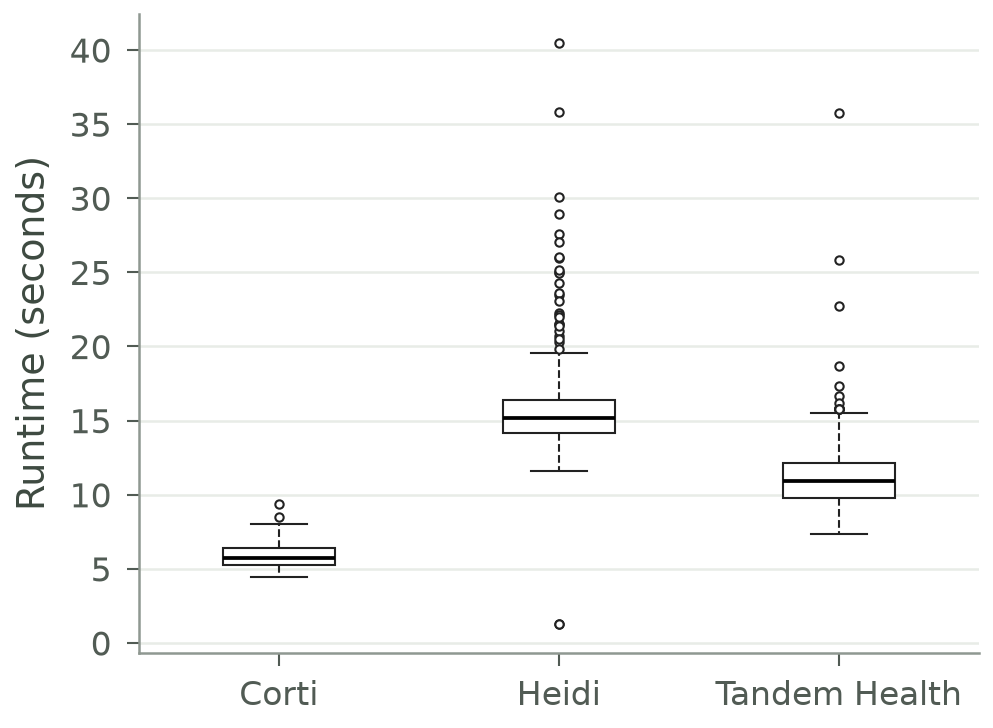}
    \hfill
    \includegraphics[width=0.48\linewidth]{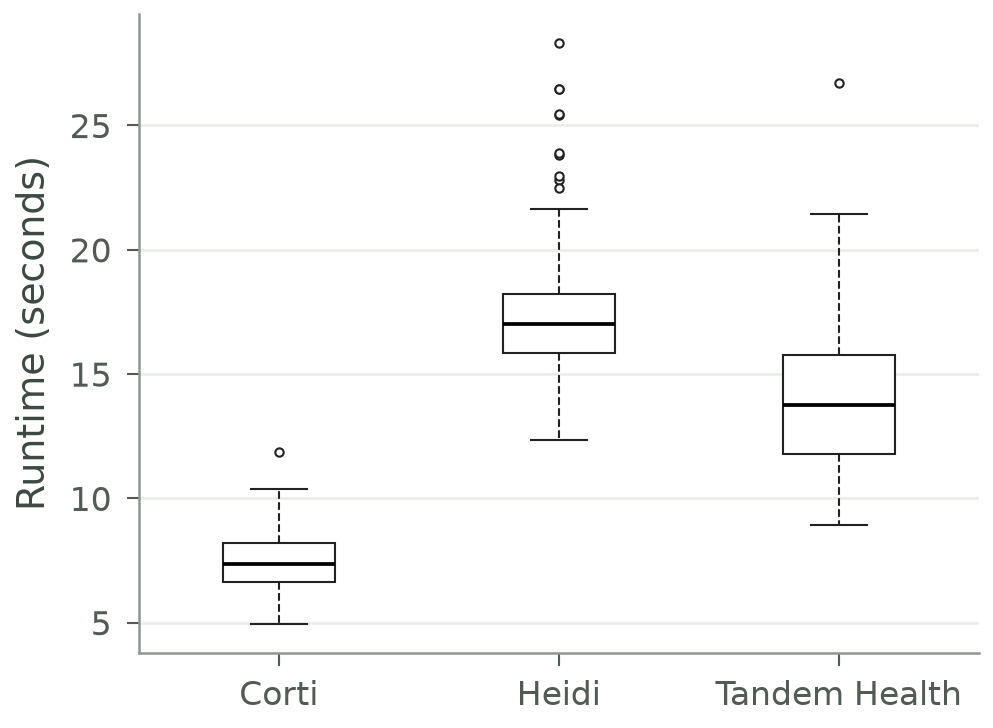}

    \medskip

    \includegraphics[width=0.48\linewidth]{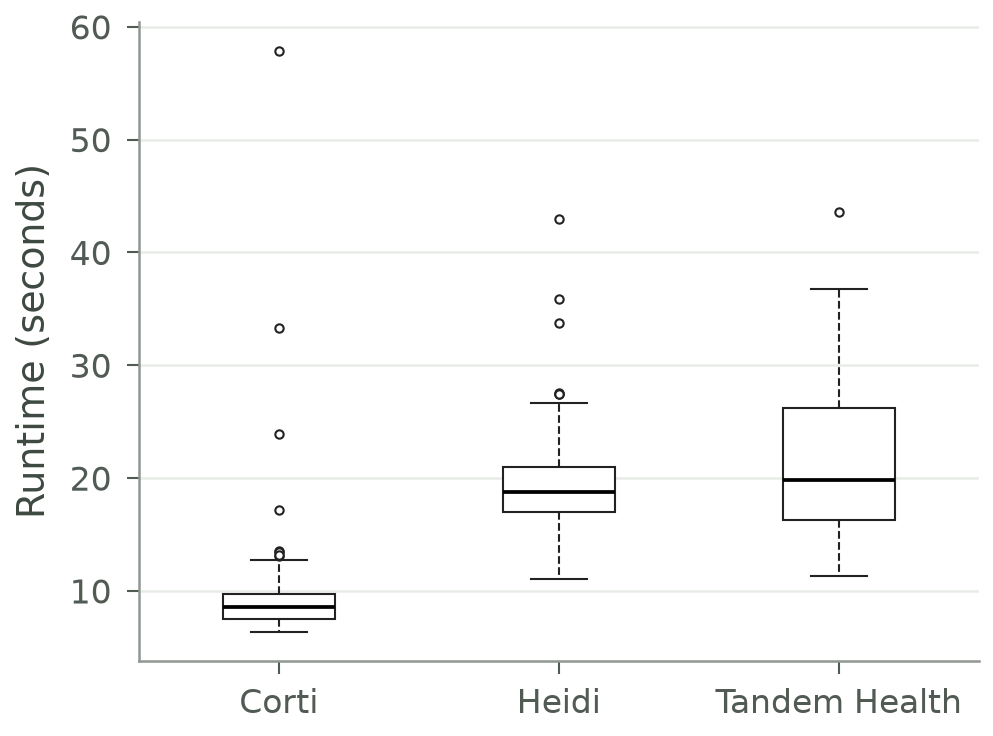}
    \hfill
    \includegraphics[width=0.48\linewidth]{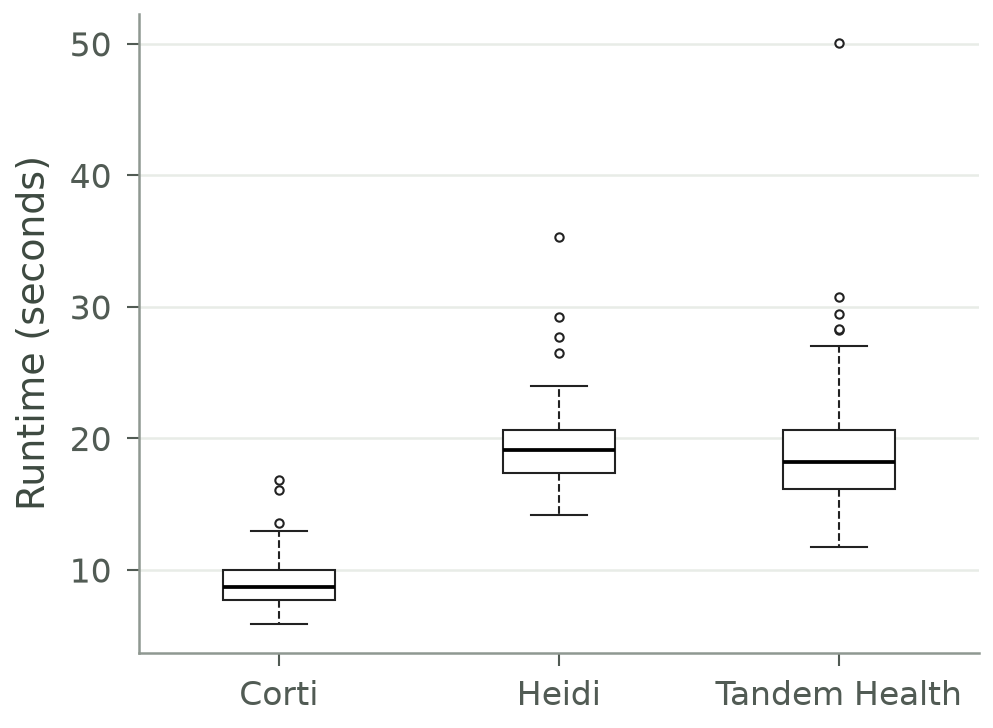}
\end{minipage}
\caption{Comparison of the latency speed plots for ACI-BENCH (top left), MedConv EN (top right), DA (bottom left) and DE (bottom right).}
\label{fig:latency-boxplots}
\end{figure}

~\autoref{fig:latency-boxplots} presents note-generation times for each system across the four datasets. Corti displays the fastest runtime across systems and datasets, with lower median, fewer outliers, and narrower interquartile range. These distributions complement the mean runtimes in \autoref{tab:latency}. We emphasize that Heidi and Tandem Health were measured through their browser interfaces, therefore, distributions should be interpreted as indicative

\section{Targeted clinician validation of the pairwise judge}
\label{appendix:judge-validation}
\begin{table}[htbp]
\centering
\caption{Inter-clinician agreement with the LLM judge across PDSQI dimensions.}
\label{tab:pdsqi_reliability}

\begin{tabular}{lrrrr}
\toprule
\textbf{Dimension} &
\textbf{Agreement} &
\textbf{Pairwise agreement} &
\textbf{Krippendorff's $\alpha$} &
\textbf{Gwet's AC1} \\
\midrule
Overall          & 207/240 (86.3\%) & 0.767 & 0.020  & 0.694 \\
Accurate         & 29/30 (96.7\%)   & 0.933 & 0.000  & 0.929 \\
Thorough         & 27/30 (90.0\%)   & 0.867 & 0.284  & 0.837 \\
Useful           & 26/30 (86.7\%)   & 0.867 & 0.442  & 0.827 \\
Organized        & 24/30 (80.0\%)   & 0.600 & -0.208 & 0.412 \\
Comprehensible   & 26/30 (86.7\%)   & 0.733 & -0.115 & 0.653 \\
Succinct         & 24/30 (80.0\%)   & 0.667 & -0.007 & 0.510 \\
Synthesized      & 21/30 (70.0\%)   & 0.467 & -0.228 & 0.080 \\
Stigmatizing     & 30/30 (100.0\%)  & 1.000 & --     & 1.000 \\
\bottomrule
\end{tabular}

\vspace{0.4em}
\begin{minipage}{0.96\linewidth}
\footnotesize
\textit{Note.} Agreement denotes the proportion of clinician evaluations
that agreed with the LLM judgment. Krippendorff's $\alpha$ and Gwet's AC1
were calculated on the binary agree/disagree ratings. Krippendorff's
$\alpha$ was undefined for the stigmatizing-language dimension because
all ratings were identical.
\end{minipage}
\end{table}

Clinicians agreed with the LLM judging, when shown its verdict and reasoning, in most assessments (207 of 240 samples, i.e. 86.3\%), with agreement strongest on accuracy, thoroughness, and usefulness and weakest on synthesis (\autoref{tab:pdsqi_reliability}). At the case-dimension level, a majority of clinicians supported the LLM judgment in 75 of 80 comparisons (93.8\%), including unanimous agreement in 52 comparisons (65.0\%).

Overall Gwet's AC1 was 0.694 indicating substantial overall agreement. Dimension-level AC1 was highest for accuracy (0.929), thoroughness (0.837), and usefulness (0.827), and lowest for synthesis (0.080). Nominal Krippendorff's alpha was 0.020 overall, consistent with the strong imbalance in the binary agree/disagree response distribution.

Qualitative post-annotation interviews highlighted that the annotation task and UI was very clear overall, with dimensions varying in difficulty to assess. All clinicians stated they had a very clear favourite note for each case, however the driver of this clear preference differed between a clear preference for succinctness for one, a more thorough fluent sentence writing style preference for another, and a bias towards succinctness but with critical flaws in the note quickly determining a winner for the third clinician.

\FloatBarrier

\begin{figure}[h]
    \centering
    \includegraphics[width=0.8\linewidth]{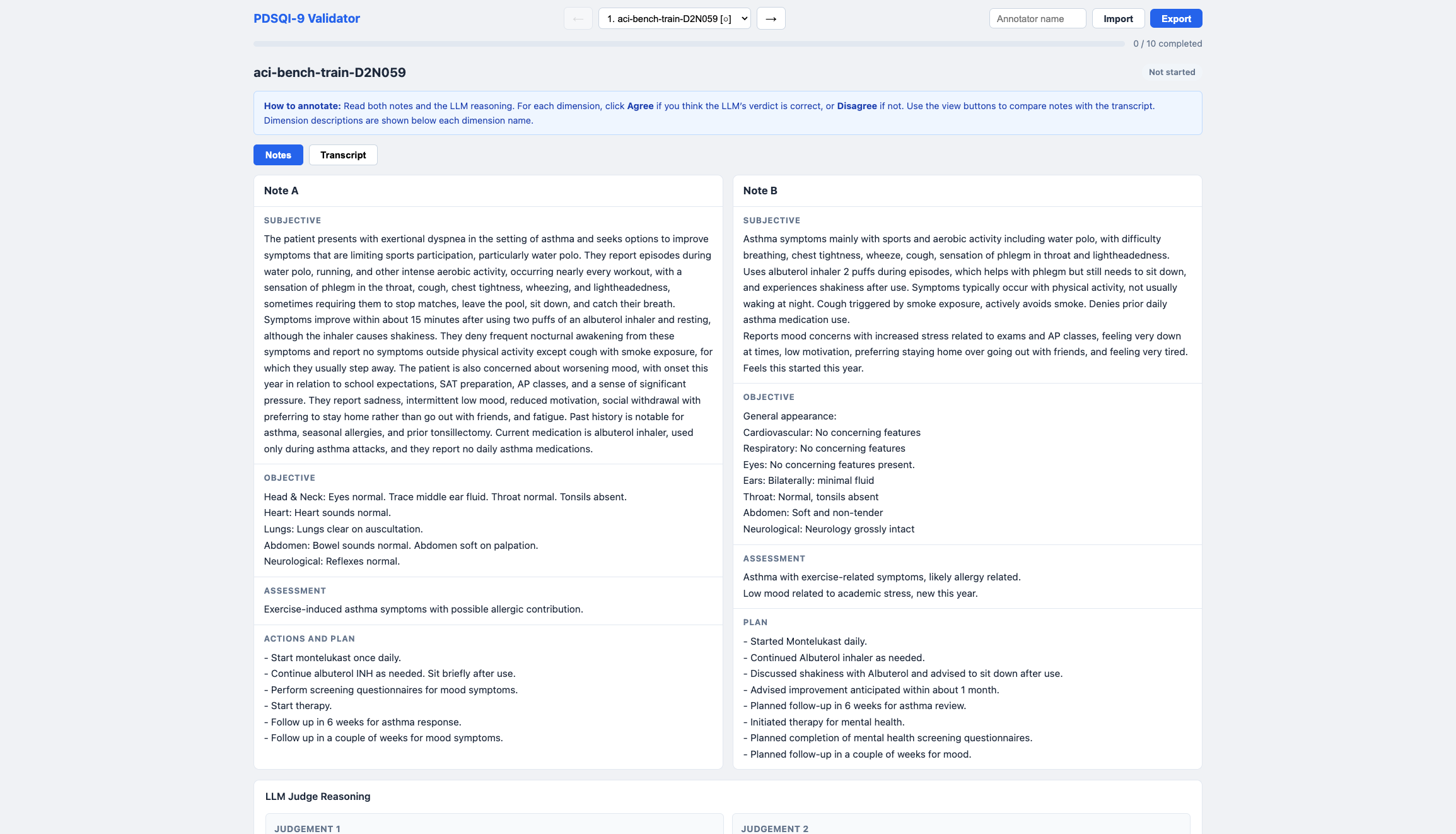}
    \caption{Validation tool showing compared system notes side-by-side, flipped randomly}
    \label{fig:pdsqi-validation-ui-1}
\end{figure}
\begin{figure}[h]
    \centering
    \includegraphics[width=0.8\linewidth]{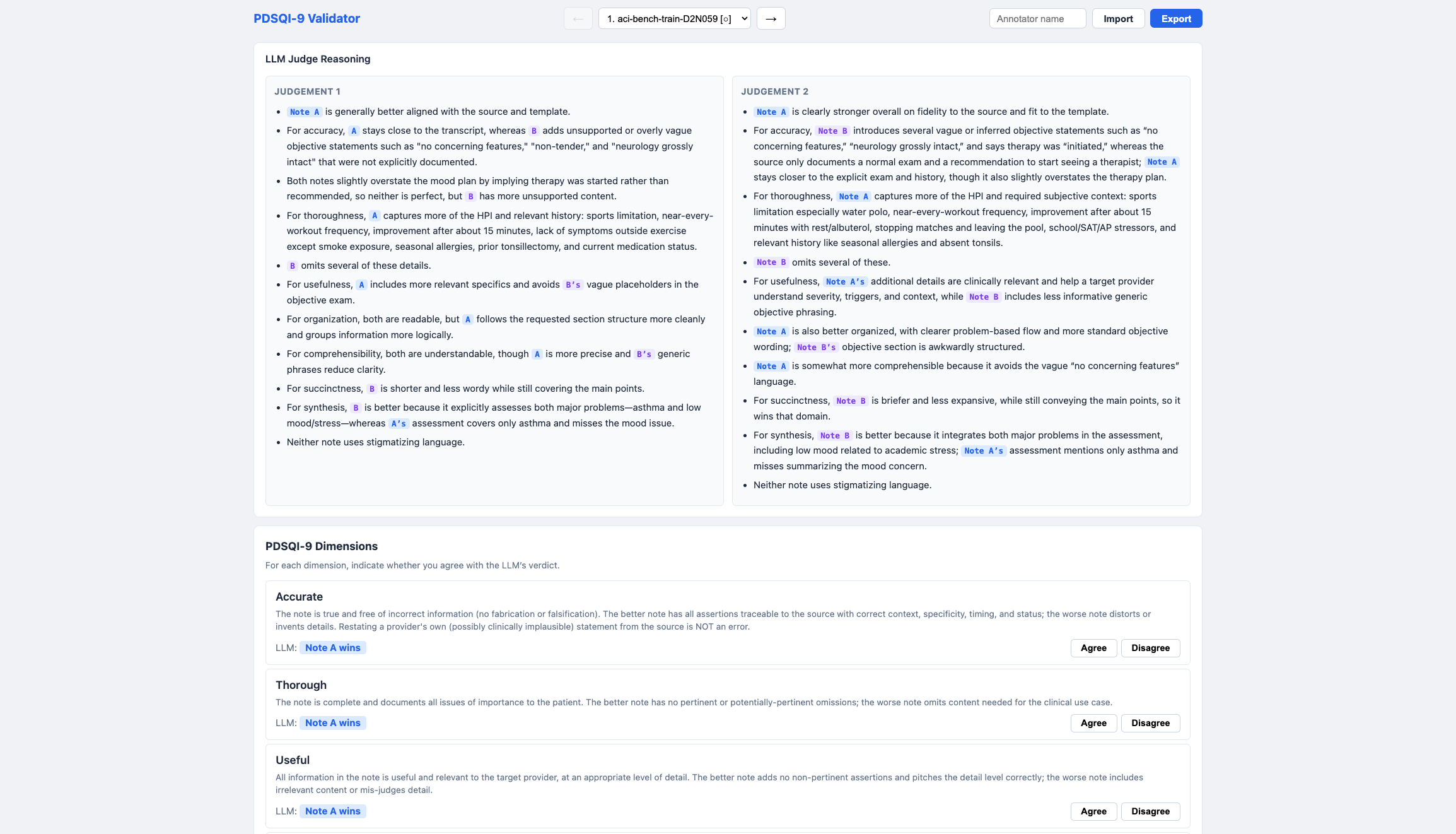}
    \caption{Validation tool showing LLM judgements and dimensions to validate}
    \label{fig:pdsqi-validation-ui-2}
\end{figure}

\section{Limitations}
\label{appendix:limitations}
The following limitations should be noted.

\begin{enumerate}

\item \textbf{Template prompt transparency.} Corti exposes all section-level prompts fully. Heidi exposes the underlying prompt in a Structure-view. Tandem Health does not expose its section-level prompts; templates created from headings only are assumed to be outfitted with hidden Tandem Health-authored prompts. This asymmetry means that for Tandem Health, the effective prompt is not fully known and cannot be directly compared to the prompts used for Corti and Heidi. Differences in output quality may therefore reflect differences in hidden prompt engineering rather than in pipeline or model capability, and this could favour Tandem Health as easily as it disadvantages it. Tandem Health also had no suitable vendor-provided UK template, so the English comparisons evaluate it under a template we built from section headings alone.

\item \textbf{Model and infrastructure differences.} The three systems use different underlying LLMs and cloud infrastructure. Corti uses its own models accessible via API. Heidi's subprocessors indicate GCP hosts ``many of our AI models'' alongside AWS and Azure. Tandem Health's subprocessor list names Microsoft Azure for ``AI models and related processing services'' and ElevenLabs for speech-to-text. The systems' LLMs, model versions, and inference configurations are not documented and may change over time. Model, prompt, template, and serving infrastructure vary together across the three systems, so observed differences are attributable to the systems as configured and served, not to any single component.

\item \textbf{Clinician validation measured ratification, not independent preference.} Clinicians were shown the LLM judge's declared winner together with its reasoning trace and asked only to agree or disagree, rather than recording an independent preference for later comparison. This anchors the annotator on the judgment under test, so the reported agreement rate and Gwet's AC1 should be read as upper bounds on judge--clinician concordance.

\item \textbf{System-level configuration parameters.} Heidi and Tandem Health both expose user-facing settings that influence output style and length (Heidi: voice modes brief/detailed/goldilocks, Simple vs.\ Pro scribe; Tandem Health: ``Brief assessment \& Plan'' toggle, ``Level of detail'' beta with Standard/Concise/Comprehensive). These were held constant at their respective defaults, but they represent system-specific parameters that are injected together with a template. In Corti, equivalent voice, style and length controls are explicitly part of the section prompting instructions itself for full transparency and control.

\item \textbf{Language and market localisation.} Heidi's Danish templates shipped with English-language prompts despite Danish headings. Tandem Health's Danish SOAP was based on a Danish-specific template (Almen Praksis). Corti's templates were configured with per-locale variants. The degree of native-language prompt localisation varies and may affect output quality in non-English markets, particularly Danish.

\end{enumerate}

\section{Characterization of ungrounded statements}
\label{appendix:ungrounded-details}
As a preliminary analysis of groundedness failures, we used an LLM (Corti-S1\footnote{https://docs.corti.ai/models}) to categorize statements flagged as partially or not entailed and assign a severity level. Across all 412 encounters, Corti accumulated 655 flagged statements (160 at severity S3+), compared with 1{,}242 (534 S3+) for Tandem Health and 1{,}286 (459 S3+) for Heidi. Fully unsupported statements were least frequent for Corti (23), followed by Heidi (83) and Tandem Health (149); most flagged content across all systems was only partially entailed, where the core claim was supported but an unstated detail had been added.

The dominant failure mode across all three systems was \emph{specificity overreach}, such as adding a medication route or laterality not present in the source. This accounted for 22\%, 26\%, and 40\% of flagged entries for Corti, Tandem Health, and Heidi, respectively. Many of these cases were low severity because the added detail was plausibly correct—for example, adding PO for a medication available only orally—but still failed strict entailment because the detail was not explicitly supported by the transcript. System-specific patterns also emerged: Tandem Health had the highest rate of \emph{fabrication from silence} (195 entries, versus 25 for Corti and 118 for Heidi), while Heidi showed more \emph{terminology substitution} (156 entries, versus 55 for Corti).

Because both failure-mode classification and severity scoring were LLM-assigned and the rubric has not been independently clinically validated, this analysis should be interpreted as an indicative characterization of the errors captured by the groundedness metric rather than a confirmed clinical error audit.

\FloatBarrier

\begin{figure}[H]
    \centering
    \caption{Flagged Statements, partially or fully ungrounded, by Dataset x Severity}
    \includegraphics[width=1\linewidth]{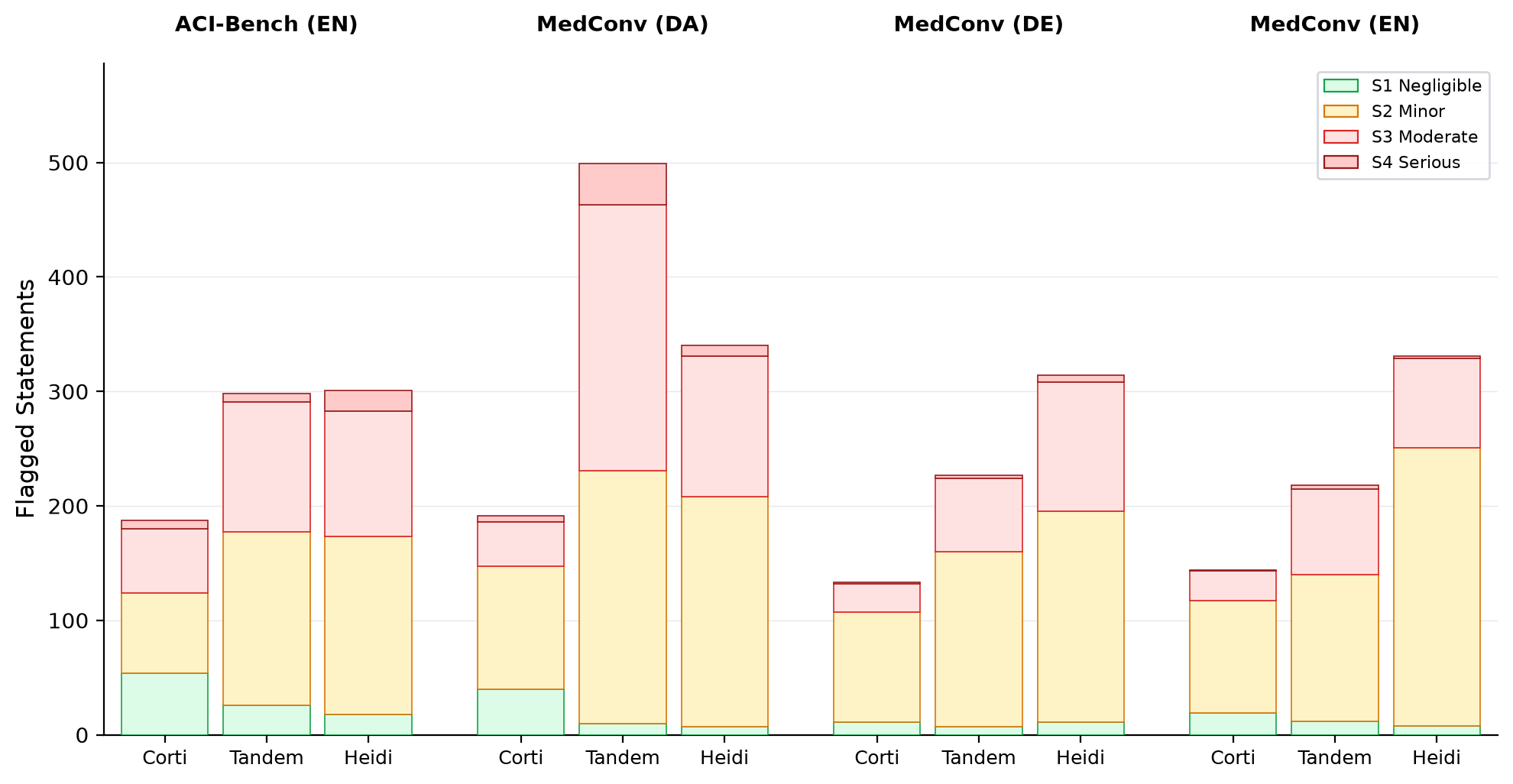}
    \label{fig:ungrounded-statement-details}
\end{figure}

{\footnotesize
\renewcommand{\arraystretch}{1.1}
% Bold severity label; encounter ID set smaller and in gray below it
\newcommand{\sevid}[2]{\textbf{#1}\newline{\scriptsize\color{gray}#2}}
\begin{longtable}{@{}
    >{\raggedright\arraybackslash}p{0.11\textwidth}
    >{\raggedright\arraybackslash}p{0.30\textwidth}
    >{\raggedright\arraybackslash}p{\dimexpr0.59\textwidth-4\tabcolsep\relax}
@{}}
\caption{Examples of ungrounded statements by system and severity (S1 negligible, S2 minor, S3 moderate, S4 severe). All examples are drawn from ACI-Bench; encounter IDs omit the \texttt{aci-bench-} prefix. Justifications are abridged from the LLM classifier output.}
\label{tab:ungrounded-examples}\\
\toprule
\textbf{Sev.} & \textbf{Flagged statement} & \textbf{Category and justification} \\
\midrule
\endfirsthead
\toprule
\textbf{Sev.} & \textbf{Flagged statement} & \textbf{Category and justification} \\
\midrule
\endhead
\midrule
\multicolumn{3}{r@{}}{\textit{Continued on next page}} \\
\endfoot
\bottomrule
\endlastfoot

% ---------------------------------------------------------------- Corti
\multicolumn{3}{@{}l}{\textbf{Corti}} \\
\cmidrule{1-3}
\sevid{S1}{test1-D2N108} &
Rx clindamycin 400 mg PO twice daily for 7 days. &
\emph{Added route / frequency.} The clinician said ``you take that,'' implying oral administration; adding ``PO'' would not mislead a reader. \\

\sevid{S2}{test1-D2N107} &
Rx physical therapy with home stretches and exercises. &
\emph{Specificity overreach.} The source supports PT with stretches and exercises but does not specify ``home.'' Clinically implied and unlikely to change management. \\

\sevid{S3}{test1-D2N109} &
Air splint applied. Crutches provided. Non-weightbearing until radiograph result. &
\emph{Contingent to definitive.} The clinician describes a plan (``I'm gonna put you in an air splint''), but the note records it as completed, misrepresenting what was done during the visit. \\

\sevid{S4}{valid-D2N077} &
Ulnar styloid fracture present. &
\emph{Fabrication from silence} The source mentions ulnar styloid fracture only in a general description of Colles' fractures, not as a finding for this patient. The note asserts a fracture that was not identified. \\

\addlinespace[2pt]
% ---------------------------------------------------------------- Tandem Health
\multicolumn{3}{@{}l}{\textbf{Tandem Health}} \\
\cmidrule{1-3}
\sevid{S1}{test1-D2N106} &
Start Montelukast 10 mg, 1 tablet once per day. &
\emph{Specificity overreach.} ``1 tablet'' is implied by a 10 mg prescription but is more specific than the source; it would not change management. \\

\sevid{S2}{valid-D2N081} &
Denies mucus, shortness of breath, nausea, vomiting, sensation of food getting stuck. &
\emph{Specificity overreach.} The patient's answer ``more dry'' supports a dry cough but is not an explicit denial of mucus. \\

\sevid{S3}{test1-D2N123} &
Thyroid: Thyroid gland symmetric, non-enlarged, smooth, without nodules or tenderness. &
\emph{Specificity overreach.} The source states only that the thyroid ``feels normal''; the listed findings are not stated and overstate the thoroughness of the exam. \\

\sevid{S4}{train-D2N038} &
No hepatosplenomegaly. &
\emph{Temporal distortion.} The clinician corrects the initial ``no hepatosplenomegaly'' to ``splenomegaly''; the note retains the superseded negative finding. \\

\sevid{S4}{train-D2N047} &
No chest pain or shortness of breath. &
\emph{Fabrication from silence.} The clinician asks about these symptoms but the patient never answers; the negatives are unsupported in a patient with known coronary artery disease. \\

\addlinespace[2pt]
% ---------------------------------------------------------------- Heidi
\multicolumn{3}{@{}l}{\textbf{Heidi}} \\
\cmidrule{1-3}
\sevid{S1}{test1-D2N111} &
Continue Lisinopril 20 mg daily. &
\emph{Added route / frequency.} Lisinopril is dosed once daily by convention; adding ``daily'' would not mislead a clinician. \\

\sevid{S2}{test1-D2N109} &
Experienced immediate soreness, swelling, and pain, making it difficult to walk. &
\emph{Specificity overreach.} ``Immediate'' onset is not stated in the source. Plausible, but adds unsupported temporal detail. \\

\sevid{S3}{test1-D2N110} &
Right foot examination: 1$\times$2 cm circular wound on the dorsal aspect of the lateral right foot, proximal to the fifth MT. &
\emph{Wording mismatch.} The source gives 1$\times$2 inches (about 2.5$\times$5 cm); the note's unit change misstates the wound size. \\

\sevid{S3}{test1-D2N112} &
Lower back pain for one month, chronic history of back pain for 30 years. &
\emph{Specificity overreach.} The source mentions a fall 30 years ago and that the back ``has been bothering me for a long time,'' not a continuous 30-year history. \\

\sevid{S4}{test1-D2N119} &
Investigations planned: None. &
\emph{Fabrication from silence.} The source describes reviewing pulmonary function tests and planning an asthma action plan; the note asserts an absence of workup. \\

\sevid{S4}{test3-D2N187} &
No PE changes in the fovea. &
\emph{Other.} The source states ``or pe changes in the fovea'' for the left eye, indicating RPE changes are present; the note asserts the opposite. \\

\end{longtable}
}

\end{document}